# $Ga_2O_3$ TCAD Mobility Parameter Calibration using Simulation Augmented Machine Learning with Physics-Informed Neural Network

Edric Khai Jieh Ong, Le Minh Long Nguyen, Matthew Eng, Yuhao Zhang, *Senior Member, IEEE* and Hiu Yung Wong, *Senior Member, IEEE*

***Abstract*—In this paper, we demonstrate the feasibility of performing automatic Technology Computer-Aided-Design (TCAD) parameter calibration and extraction using machine learning, with the machine *trained solely by TCAD-simulation data*. The methodology is validated using *experimental data*. Schottky Barrier Diodes (SBDs) with different effective anode workfunction (*WF*) are fabricated with emerging ultra-wide-bandgap material, Gallium Oxide ($Ga_2O_3$), and are measured at various temperatures (*T*). Their current-voltage curves are used for automatic $Ga_2O_3$ Philips Unified Mobility (PhuMob) model parameters calibration. Five critical PhuMob parameters ($\mu_{max}$, $\mu_{min}$, $N_{ref}$, $\alpha$, and $\theta$) were calibrated. The machine consists of an autoencoder and a neural network and is trained *solely* by TCAD simulation data with variations in *WF*, *T*, and the five PhuMob parameters (seven variations in total). Then, $Ga_2O_3$ PhuMob parameters are extracted from the noisy experimental curves. Subsequent TCAD simulation using the extracted parameters shows that the quality of the parameters is as good as an expert's calibration at the pre-turned-on regime, but not in the on-state regime. By using a simple physics-informed neural network, the machine performs as well as the human expert in all regimes.**

## I. Introduction

TCAD parameter calibration is critical for emerging material and device development to enable early-stage exploration when experimental study is still impossible or expensive [1]-[3]. Gallium Oxide ($Ga_2O_3$) device, which is promising for building high-voltage and high-power circuits, is such an example. However, calibration of TCAD parameters can be tedious and repetitive. For example, in [4], to calibrate the Philips Unified Mobility (PhuMob) model [5] for $Ga_2O_3$, many expensive engineering hours were required for a TCAD expert using significant amount of domain expertise. Due to uncertainties in emerging devices (such as the existence of unknown defects) and the lack of fundamental calibration structures and experimental data, the same model needs to be recalibrated for different devices. For example, $Ga_2O_3$ FinFET [4] and Schottky barrier diode (SBD) [6] are calibrated to have different $\mu_{max}$ in the PhuMob model [7]. Therefore, it is desirable to develop an automatic framework using machine learning (ML) to calibrate TCAD parameters without the need for much domain expertise. Moreover, due to the lack of experimental data, it is desirable to train the machine using TCAD simulation data (simulation-augmented ML), which has been used for transistor inverse design and surrogate model development to predict electrical characteristics from device structure parameters [8]-[14]. In some cases, TCAD is used together with optimization algorithms such as Bayesian Optimization to optimize a design [15][16] or optimization is performed with TCAD-trained neural networks [17][18].

TCAD model parameter calibration and extraction are equivalent to reverse engineering for structural parameters and measurement conditions. We have demonstrated the extraction of measurement conditions (*T*) and structural parameters (e.g., *WF*) of $Ga_2O_3$ SBD in [6] using machines trained by TCAD data. In [19] and [20], high accuracy was obtained through the application of noise and a modified autoencoder (AE). In all of these studies, manually calibrated TCAD parameters were used.

In [21], we proposed to use ML to perform automatic $Ga_2O_3$ PhuMob parameters extraction. In this paper, we show that it is possible to use TCAD to generate data to train a machine that can correlate $Ga_2O_3$ SBD current-voltage (IV) curves to PhuMob parameters ($\mu_{max}$, $\mu_{min}$, $N_{ref}$, $\alpha$, and $\theta$), device temperature, $T$, and an effective anode gate work function, $WF$. By feeding experimental data to the machine, it is able to calibrate TCAD parameters automatically to reproduce the experimental data with high accuracy. However, a physics-informed neural network (PINN) needs to be incorporated to achieve the highest accuracy. Compared to [21], this work 1) extracts the parameters based on noisy experimental data (instead of unseen simulation data), 2) requires less than one-third of the TCAD data in [21], and 3) includes $WF$ variation (first time to have 7 variations). Furthermore, this study reveals the need to use PINN and certain domain expertise when it is applied to realistic experimental data.

## II. Experiment and TCAD Model

Sixty-six $Ga_2O_3$ SBDs were fabricated (Fig. 1). The details

This material is based upon work supported by the NSF under Grant No. 2046220. Nguyen, Ong, Eng, and Wong are affiliated with San Jose State University, San Jose, CA 95192 USA. Nguyen and Ong have equal contributions. Zhang is affiliated with the Department of Electrical and Electronic Engineering, University of Hong Kong (HKU), Hong Kong. (Corresponding author: Wong; e-mail: hiuyung.wong@sjsu.edu).

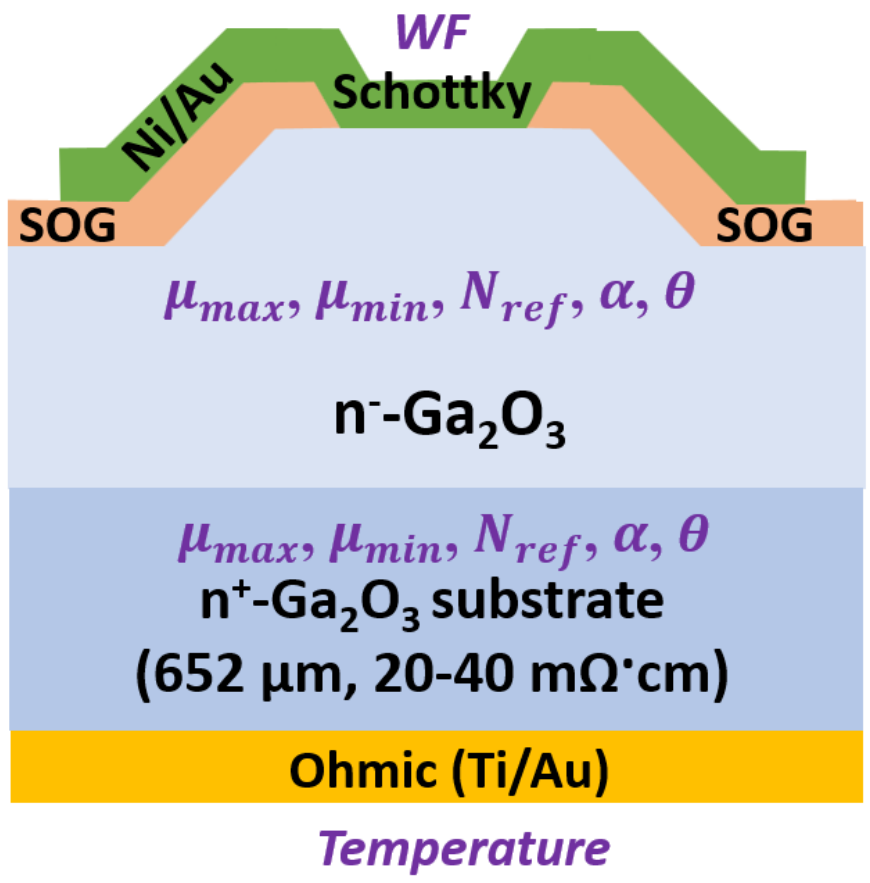

Fig. 1. Schematic of the fabricated vertical $Ga_2O_3$ power Schottky barrier diodes. The seven parameters (in purple) are shown and placed in the regions where they are expected to affect the electrical properties. Note that the PhuMob parameters are doping and carrier concentration dependent and affect both the drift and the substrate regions.

may be found in [19]. Special processing techniques were employed to create variations in the effective $WF$ of the anode metal. The devices were measured at three different temperature groups, Temp1 (298K–308K), Temp2 (348K–368K), and Temp3 (403K–423 K). The drift layer thickness and doping are also expected to have 25% to 35% variations [19]. Some IV curves are also noisy as seen in Fig. 2. *The goal is to perform PhuMob calibration using the non-ideal data of this emerging material, assisted with TCAD-generated data and PINN.* Fig. 1 shows the structure with the 7 parameters highlighted in purple. The 5 PhuMob parameters, $\mu_{max}$, $\mu_{min}$, $N_{ref}$, $\alpha$, and $\theta$, impact the carrier mobility, $\mu$, through the following equations [7][22].

$$\frac{1}{\mu} = \frac{1}{\mu_L} + \frac{1}{\mu_{DA}} \tag{1}$$

$$\mu_L = \mu_{max}\left(\frac{T}{300}\right)^{-\theta} \tag{2}$$

$$\mu_{DAeh} = \mu_N\left(\frac{N_{ref}}{6.0\times10^{15}}\right)^{\alpha} + \mu_C \tag{3}$$

$$\mu_N = \frac{{\mu_{max}}^2}{\mu_{max}-\mu_{min}}\left(\frac{T}{300}\right)^{3\alpha-1.5} \tag{4}$$

$$\mu_c = \frac{\mu_{max}\mu_{min}}{\mu_{max}-\mu_{min}}\left(\frac{300}{T}\right)^{0.5} \tag{5}$$

where $n$ and $p$ are the electron and hole densities. They affect both the drift and the substrate region and thus the IV on-state current.

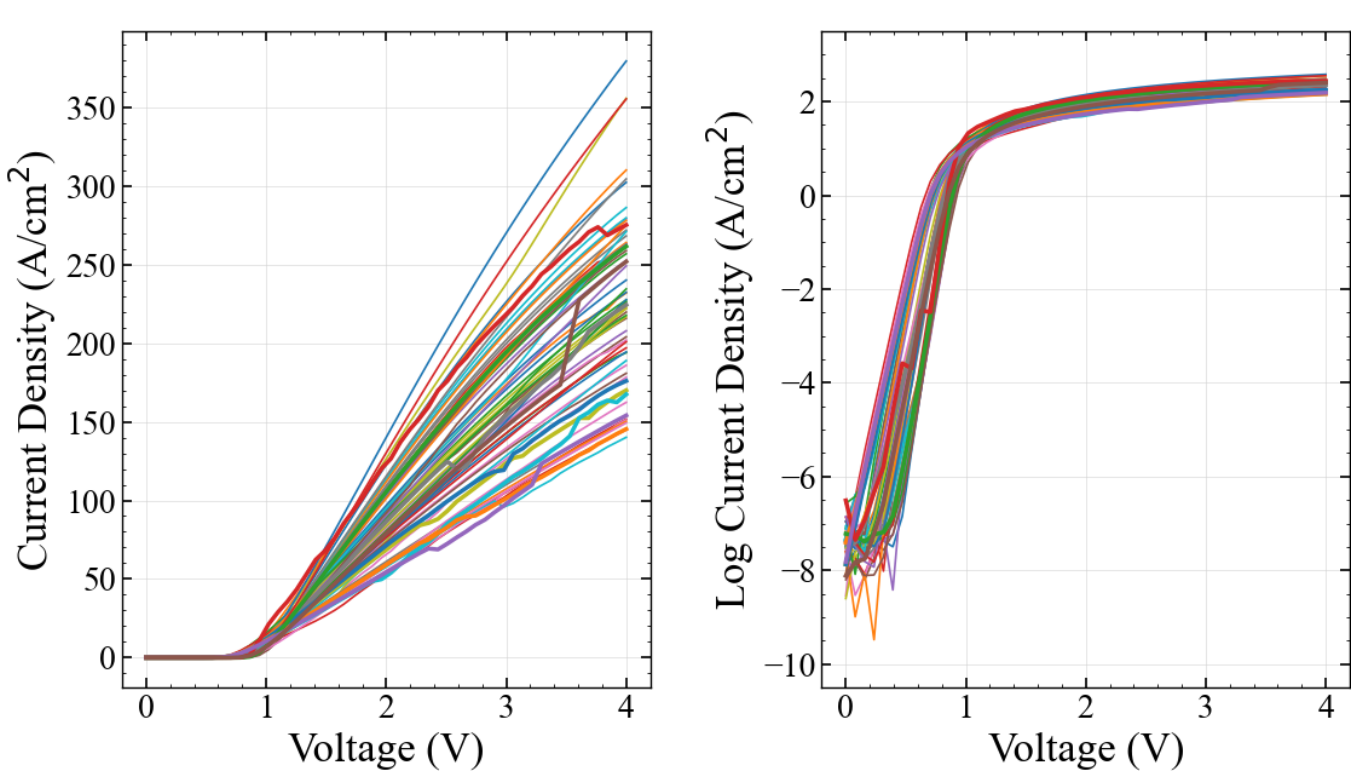

Fig. 2. 66 $Ga_2O_3$ SBD experimental I-V Curves in linear and logarithmic scales. Currents are normalized to anode area (radius = 120µm).

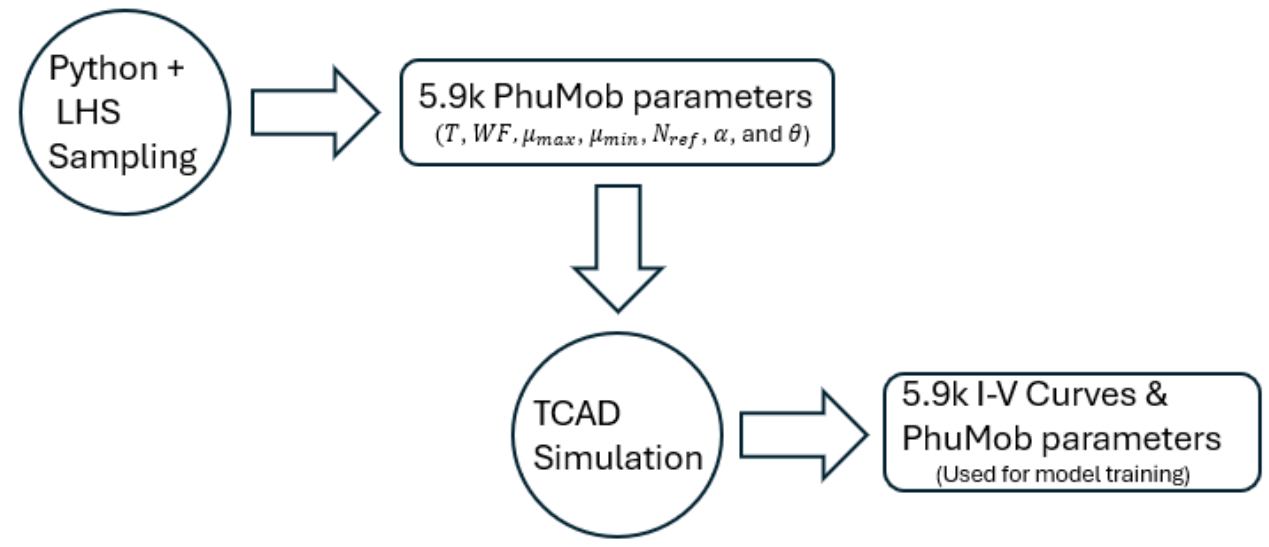

Fig. 3. Workflow illustrating the generation of 5.9 K I–V curves and the corresponding PhuMob parameters via TCAD simulation.

## III. Data Generation and Physics

TCAD simulations were used to generate ~5.9k forward IVs (0V to 4V) of a $Ga_2O_3$ SBD by varying the aforementioned 7 parameters ($T$, $WF$, $\mu_{max}$, $\mu_{min}$, $N_{ref}$, $\alpha$, and $\theta$) using the setup in [21]. TCAD Sentaurus was used [22]. Specifically, device structures were generated using Sentaurus Structure Editor, electrical simulations were performed using Sentaurus Device, and post-processing/IV data extraction was carried out using Sentaurus Visual. The whole process from structure generation to IV extraction was less than 10 minutes for each device. It took less than 10 hours to complete the full dataset generation on a 75-core shared machine. We will discuss the choice of 5.9k in Section VI. The ranges of the parameters are $T \in [200K, 500K]$, $WF \in [5.0eV, 5.5eV]$, $\mu_{max} \in [22cm^2/Vs, 2000cm^2/Vs]$, $\mu_{min} \in [20cm^2/Vs, 1810cm^2/Vs]$, $N_{ref} \in [10^{17}cm^{-3}, 10^{18}cm^{-3}]$, $\alpha \in [1,5]$, and $\theta \in [0.5,5]$. The Latin Hypercube Sampling (LHS) technique is used to generate the random parameter values for each TCAD simulation. LHS is used because it can enhance parameter space coverage [21][23]. Fig. 3 shows the data generation process. Each curve is discretized to 52 points. However, the first point (V = 0) was dropped as the current at V = 0 should not be affected by the PhuMob parameters (supposed to be I = 0A). By dropping the first point, better results were obtained. Therefore, 51 points are used. 72%, 13%, and 15% of the data are used for training, validating, and testing, respectively, which have been shown to be most optimized for parameter calibration in previous work [21] and are within the typical range used in most machine learning works.

Two pieces of domain knowledge are used in data generation. Firstly, a constrained sampling scheme [21] is used to generate data with $\mu_{max} > \mu_{min}$ by sampling the difference between $\mu_{max}$ and $\mu_{min}$. Secondly, $N_{ref} > 10^{17}cm^{-3}$ is used as another constraint. The impurity and carrier scattering mobilities are given by Eq. (3), where $\mu_N$ and $\mu_C$ are mobilities depending on $\mu_{max}$, $\mu_{min}$, $\alpha$, and $T$ (Eqs. (4) and (5)). When $N_{ref}$ is unphysically small, it is found that the mobility dependence on these four parameters cannot be deconvolved. Such data results in a machine that is too general and cannot predict realistic devices well. However, it should be noted that setting $N_{ref} > 10^{17}cm^{-3}$ *does not ignore* $\mu_{DAeh}$ compared to phonon scattering mobility ($\mu_L$). Otherwise, the PINN condition is to be used later (setting a penalty for $\mu_{max} < \mu_{min}$ prediction) would not have been necessary. The 7 parameters are scaled using a MinMaxScaler to [0,1]. The 51 input current points are standardized using StandardScaler, which normalizes the data to have zero mean and unit variance.

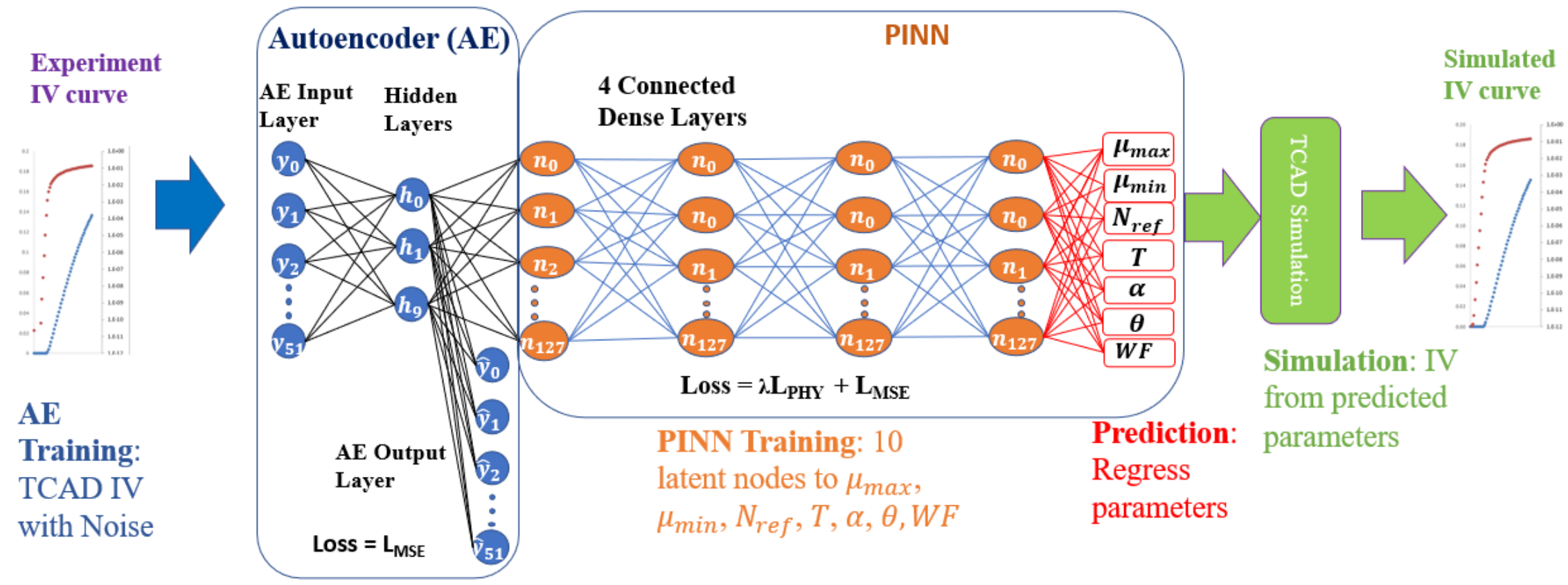

Fig. 4. Automatic calibration framework used in this flow. The AE and PINN are trained and tested using the 5.9k simulation data. To calibrate PhuMob, 66 experimental $Ga_2O_3$ SBD IV are fed from the left through the AE and PINN to obtain the parameters (one of the curves is shown in both linear and logarithmic scales). To verify the calibration, TCAD simulations are performed to obtain the IV. $R^2$ between each experimental and simulation IV pair is calculated.

## IV. Machine Learning Frameworks

Two ML frameworks are tested. The first one integrates an autoencoder (AE) and a neural network (NN), dubbed AE-NN. Another version integrates an AE with a PINN, dubbed AE-PINN (Fig. 4). The AE compresses 51-dimensional I–V curve data into a 10-dimensional latent space, effectively capturing essential device characteristics. This AE is first trained independently to minimize reconstruction loss, ensuring meaningful feature extraction, using the aforementioned TCAD-generated data. The resulting latent representations then serve as input to a dense NN or PINN. The NN or PINN features a six-layer architecture consisting of a 10-node input layer (receiving the autoencoder's latent space representations), four hidden layers with 128 nodes each, and a 7-node output layer.

*Autoencoder training:* The autoencoder takes a 51-input IV curve and feeds it into a fully connected encoder/decoder layer with node sizes of 128 - 64 - 32 - 10 (latent) - 32 - 64 - 128 - 51, using ReLU activation functions in all hidden layers and a linear activation function at the output layer. The AE was trained using MSE loss with the default TensorFlow Adam optimizer parameters. It has a custom learning rate scheduler that reduces as epoch increases (details below) and batch size of 256. It uses early stopping with patience of 500 and restores weights of the best epoch with the lowest validation loss. The autoencoder training stopped early at 2865 epochs with max epoch set to 50000. To improve robustness and reduce overfitting [20], we also added Gaussian noise at 35dB SNR to the inputs prior to normalization (Fig. 5) while clean normalized IV curves are set as the targets (Figs. 4 and 5). In total, there are 4255 training samples, 752 validation samples, and 884 test samples where we use a standard machine learning random split of ratio (72/13/15%) with a set seed of 5. Inputs and targets were normalized using a standard scalar with zero mean and unit variance. We use a piecewise-constant learning rate schedule starting at $1\times10^{-4}$ for the first 50 epochs, increasing to $7\times10^{-4}$ for epochs 51–100, then gradually decaying back to $1\times10^{-4}$ beyond epoch 2500.

*NN/PINN training*: As for the NN architecture, clean IVs are fed into the trained AE, and its latent representation (10 nodes) is taken as input into a fully connected dense neural network with 4 hidden layers of 128 nodes each with ReLU activation function and a 7-node output layer with linear activation function. The NN was first pretrained with MSE loss only and weights were saved, which were then used to initialize both the baseline NN and PINN. Both NN/PINN use Adam optimizer with default parameters, batch size of 256, and early stopping with patience of 500. The NN training stopped early at epoch 643 with a max epoch set to 50000. The PINN shares the same architecture as NN and was trained for 698 epochs before early stopping but it incorporates a hybrid loss function that combines conventional MSE with physics-based constraints [24][25]. The physics-informed part is achieved by adding the rectified value of $\hat{\mu}_{min} - \hat{\mu}_{max}$ (predicted values) in the batch, scaled by a hyperparameter $\lambda$:

$$L_{\text{total}} = L_{\text{MSE}} + \lambda L_{\text{phy}} \quad (6)$$

$$L_{\text{MSE}} = \frac{1}{N\times M}\sum_{i=1}^{N}\sum_{j=1}^{M}\left(y_{\text{true},i,j} - y_{\text{pred},i,j}\right)^2 \quad (7)$$

$$L_{\text{phy}} = \frac{1}{N}\sum_{i=1}^{N} max(\,0, \hat{\mu}_{min,i} - \hat{\mu}_{max,i}) \quad (8)$$

where N = 256 is the batch size of samples and M = 7 is the number of parameters. $y_{\text{true},i,j}$ and $y_{\text{pred},i,j}$ are the true and predicted j-th parameter of the i-th curve, respectively. $\hat{\mu}_{min,i}$ and $\hat{\mu}_{max,i}$ are the i-th predicted values of $\hat{\mu}_{min}$ and $\hat{\mu}_{max}$ respectively. Max (0,·) is the ReLU function. Therefore, if $\hat{\mu}_{max,i} > \hat{\mu}_{min,i}$, it adds no loss. If $\hat{\mu}_{max,i} < \hat{\mu}_{min,i}$, an extra loss (penalty) is added to impose the physical requirement of $\mu_{max} > \mu_{min}$ in the training process. This enhancement addresses a critical limitation in the AE-NN model, which frequently generates physically impossible mobility parameters ($\mu_{min} > \mu_{max}$). It should be noted that $\mu_{max} > \mu_{min}$ can ensure that the mobilities in Eqs. (4) and (5) to be positive. Table I shows that the AE-PINN model significantly reduces these violations. This improvement confirms that embedding physics-based constraints significantly enhances the model's ability to maintain correct physical behavior.

TABLE I
Number of Data Points Violating $\hat{\mu}_{max} > \hat{\mu}_{min}$

| Dataset | AE-NN | AE-PINN λ = 0.05/0.96 |
|---|---|---|
| Test | 7 | 2/0 |
| Training | 13 | 2/0 |
| Experiment | 2 | 1/0 |

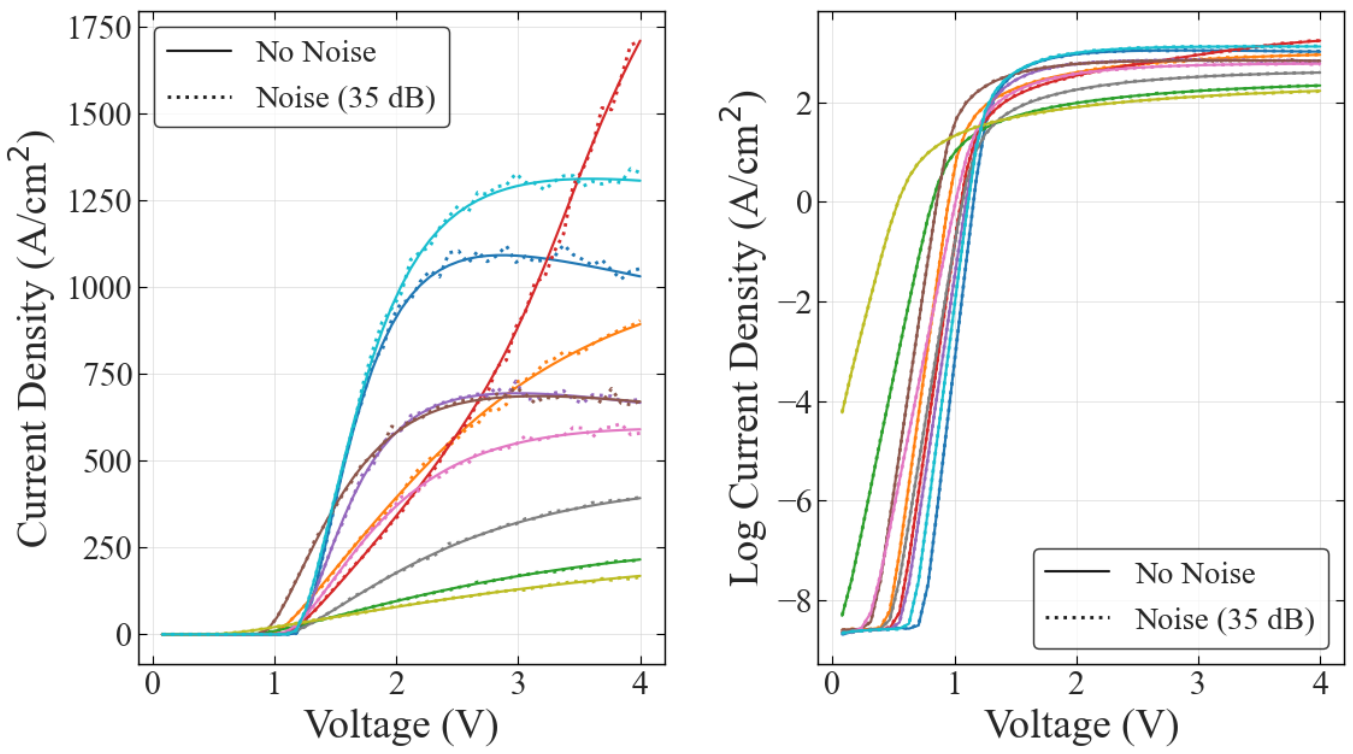

Fig. 5. TCAD generated training samples with and without noise (35dB SNR) in linear and log scales. Currents are normalized to anode area (radius = 120μm).

To optimize the hyperparameter λ, we perform a sweep of λ values to determine the optimal validation loss (Fig. 6). There are two local minima, λ = 0.05 and λ = 0.96. Both give a similar result, and λ = 0.05 is used in the following study, and we will discuss the choice of λ = 0.05 in Section VI.

Selecting an optimal λ is crucial for balancing empirical accuracy and adherence to physical constraints. A too small λ inadequately enforces physical constraints, causing the model to primarily minimize the MSE loss and neglect physics considerations. Conversely, an excessively large λ prioritizes physical constraints excessively, sacrificing empirical accuracy.

## V. Results

To perform PhuMob parameter extraction and to validate the accuracy of the AE-NN and AE-PINN ML models, the 66 experimental data are fed into the machine as shown in Fig. 4.

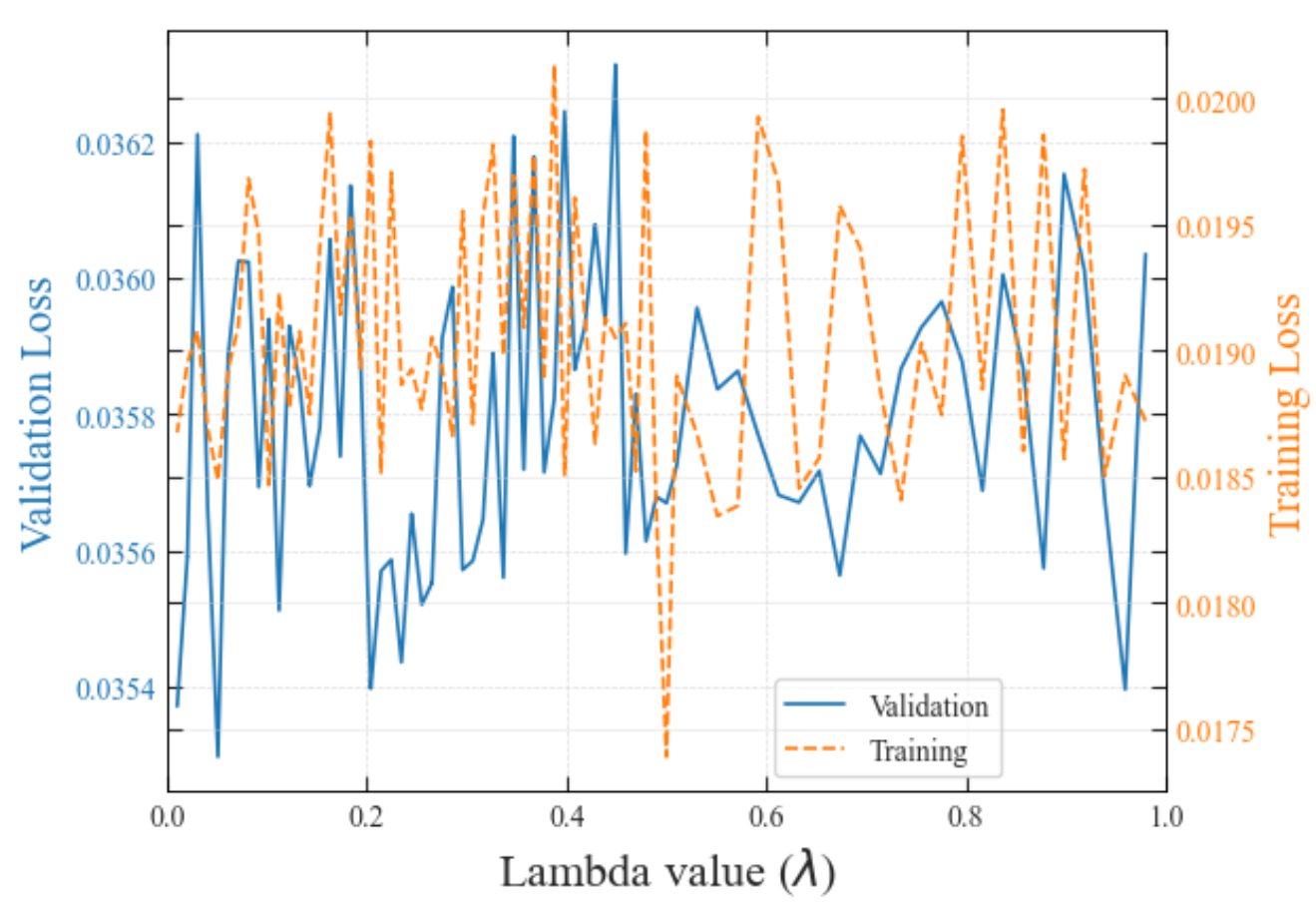

Fig. 6. Plot of validation and training losses in the AE-PINN model as a function of λ.

TABLE II
PARAMETER PREDICTIONS FOR 66 EXPERIMENTAL DATA

| Parameter | Manual | AE-NN | AE-PINN λ=0.05 |
|---|---|---|---|
| $\mu_{max}$ (cm$^2$ V$^{-1}$s$^{-1}$) | 123 | 225 | 187 |
| $\mu_{min}$ (cm$^2$ V$^{-1}$s$^{-1}$) | 80 | 75 | 60 |
| log ($N_{ref}$ (cm$^{-3}$)) | 17.3 | 17.5 | 17.5 |
| α | 0.9 | 2.8 | 2.8 |
| θ | 1.8 | 2.6 | 2.4 |

The AE extracts latent values and predicts the 7 parameters through the NN or PINN. Since all experimental data share the same epitaxial layers, which determine the I-V through mobilities, they are expected to share the same set of PhuMob parameters. The average values of the PhuMob parameters are thus used as the calibrated values.

Table II compares the extracted values to those by an expert [7], AE-NN, and AE-PINN. It can be seen that AE-PINN has outperformed AE-NN with $\mu_{max}$ and $\theta$ closer to manual calibration by an expert. To further verify the validity of the extraction, TCAD simulations are conducted using the extracted parameters with the corresponding extracted $T$ and $WF$ for each curve. Note that $T$ for each curve is the average extracted $T$ of the same temperature group. The simulated I-V is compared to the experimental one using the $R^2$ score.

Table III presents the average $R^2$ across all 66 experimental datasets, assessed on both linear and log-scaled I-V curves. It can be seen that without PINN, the prediction of on-state current (which dominates on a linear scale) is inaccurate. The PINN-based model achieves the highest $R^2$ scores, nearly matching expert-level calibration accuracy. On the other hand, with and without PINN give similar predictions in the pre-turned-on regime (dominates on a log scale). This is reasonable

TABLE III
SIMULATION IV $R^2$ (LINEAR AND LOGARITHMIC SCALES) OF DIFFERENT CALIBRATION METHODS

| | Manual | AE-NN | AE-PINN λ=0.05 |
|---|---|---|---|
| Linear | 0.944 | 0.831 | 0.933 |
| Logarithmic | 0.990 | 0.989 | 0.989 |

because the physics injected is related to the on-state mobility.

Fig. 7 reveals that both the AE-PINN approach and manual calibration methods yield not only higher $R^2$ scores but also more consistent performance, evidenced by their narrower interquartile ranges. The AE-NN approach, however, displays significantly greater variability, lower median $R^2$ values, and more outliers, highlighting its inconsistent performance compared to both the AE-PINN and manual calibrations.

To further understand how each experimental curve is matched with the three different calibration methods, Table IV summarizes the relative $R^2$ score count for AE-NN and AE-PINN against manual calibration. AE-NN and AE-PINN have similar performance on the log scale but differ in the linear scale. AE-PINN outperforms AE-NN with higher counts of calibration while producing fewer "worse" I-V curves than AE-NN. This indicates that incorporating physics-informed loss improves model performance in the linear operating regime. In Fig. 8, we show an example device I-V comparison among manual calibration, AE-PINN, AE-NN and experimental data, in which AE-PINN gives the best result.

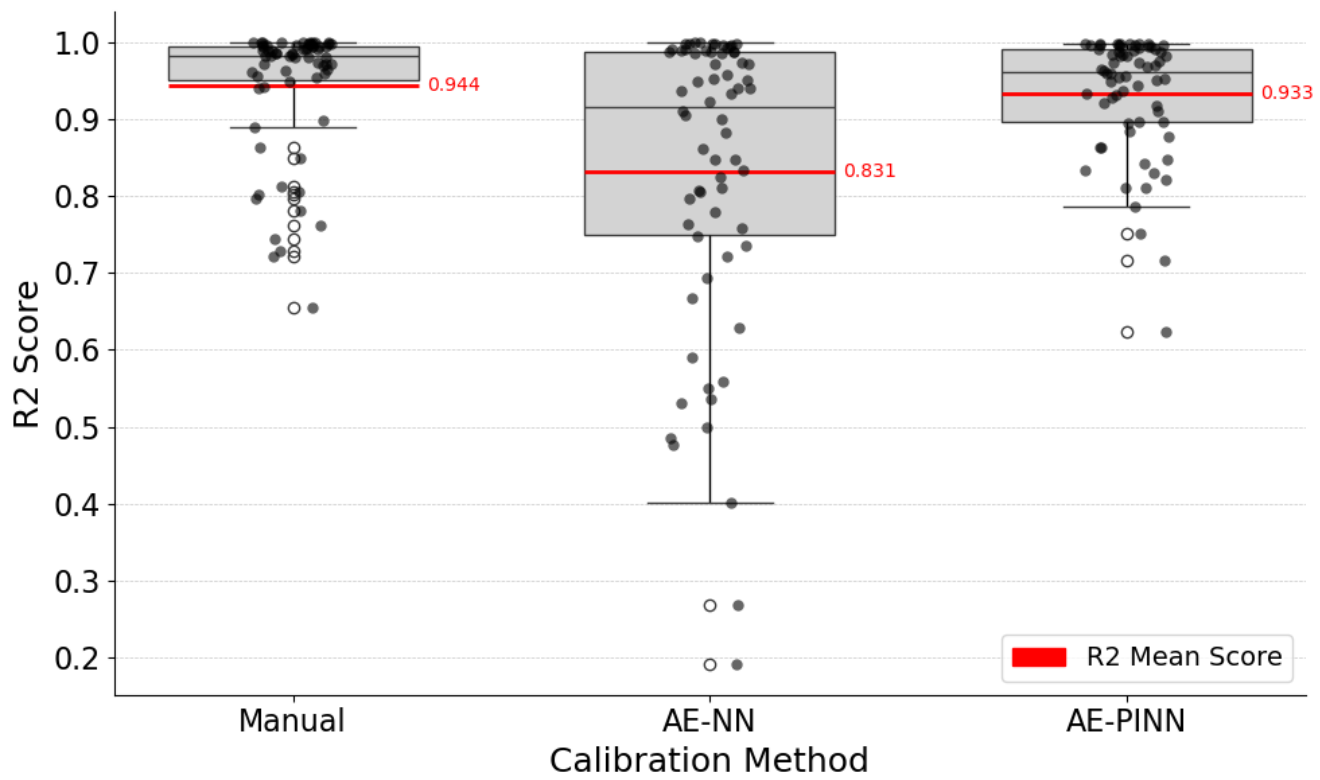

Fig. 7. Whisker box plot of the $R^2$ scores of 66 TCAD simulation I-V curves using the extracted parameters compared to the experimental I-V curves from different calibration methods - Manual, AE-NN, and AE-PINN model. Curves are in a linear scale. The AE-PINN and Manual perform better than AE-NN. AE-PINN and Manual calibration methods are similar. Note that the manual characterization is by an expert with many years of experience in TCAD.

TABLE IV
RELATIVE $R^2$ SCORE COUNTS (COMPARED WITH MANUAL CALIBRATION)

| | AE-NN(Log) | AE-PINN(Log) | AE-NN(Linear) | AE-PINN(Linear) |
|---|---|---|---|---|
| Better | 3 | 2 | 9 | 11 |
| Similar | 59 | 59 | 16 | 33 |
| Worse | 4 | 5 | 41 | 22 |

## VI. DISCUSSION

Using the proposed methodology, the process is faster than manual calibration and uses much less human engineering time (most of the time is server computation time). The major benefit of using this approach is that it does not require significant domain expertise. During the manual calibration process for this emerging material and device, a well-trained engineer with strong device physics is needed to decide which curve to use for calibration, to find the initial values for the 7 parameters, and to analyze the results (using device physics knowledge) to understand the trend to further adjust the parameters. Therefore, this process incurs significant human engineering time and effort.

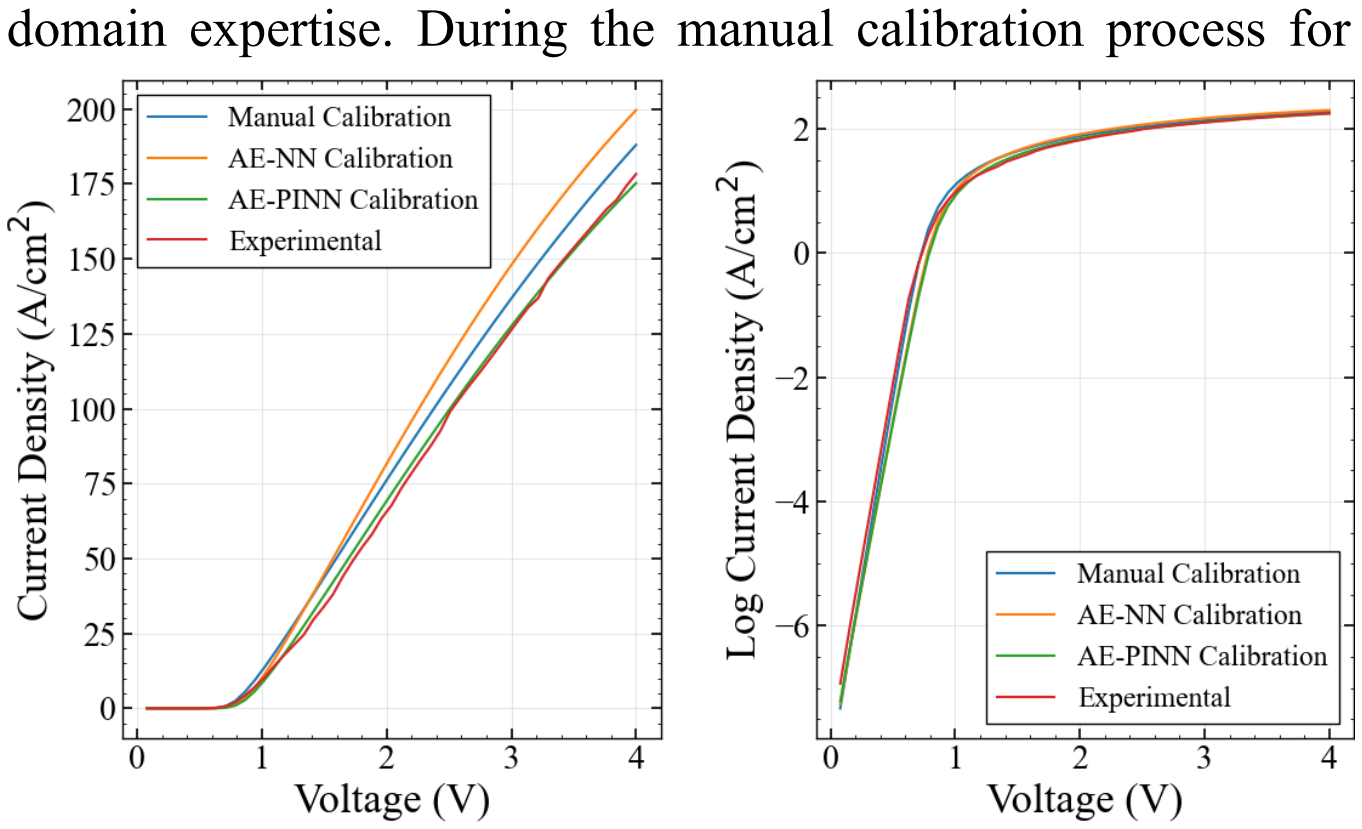

Fig. 8. Experimental I-V and TCAD I-V plots of an example device by manual, AE-NN, and AE-PINN calibration methods. Currents are normalized to anode area (radius = 120µm).

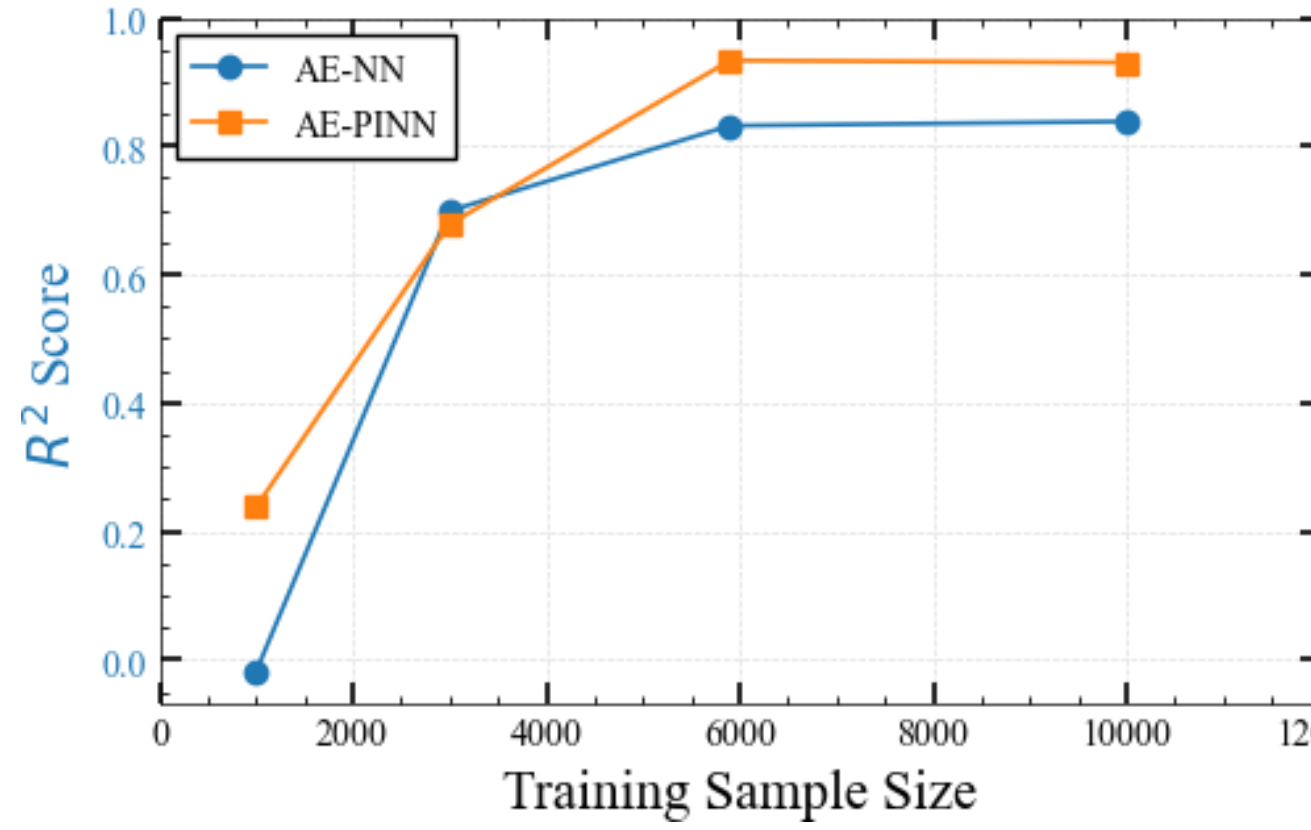

Fig. 9. Plot of $R^2$ score vs. training sample size for I–V curve prediction accuracy for AE-NN.

In order to accurately identify the minimum number of training samples sufficient to reach near-optimal accuracy and avoid unnecessary simulation overhead, we performed a study by varying the size of the training dataset and evaluating the resulting prediction accuracy. The AE-NN model was trained using subsets of 1k, 3k, 5.9k, and 10k TCAD I–V curves, and the predictions were compared against experimental data using the average $R^2$ score. As shown in Fig. 9, the $R^2$ score improves sharply between 1k and 3k but begins to saturate beyond 5.9k samples. Similarly, in Fig. 9, the AE-PINN model was trained and compared with a similar approach as AE-NN and its average $R^2$ improves linearly between 1k and 5.9k and also saturates beyond 5.9k samples. This suggests that beyond 5.9k data points, the prediction by both models do not get significantly better. Thus, we selected ~5.9k samples as the minimum dataset size for training both the AE-NN and Physics-informed (AE-PINN) models.

It should be noted that such an approach is not "a posteriori" and it does not change the effectiveness of the framework proposed. This is because TCAD data generation consumes more than 95% of the computation time in the auto-calibration framework. One may start with a small number of data generations and plot $R^2$ as in Fig. 9 and stop when it saturates. This scheme will not add too much overhead to the process.

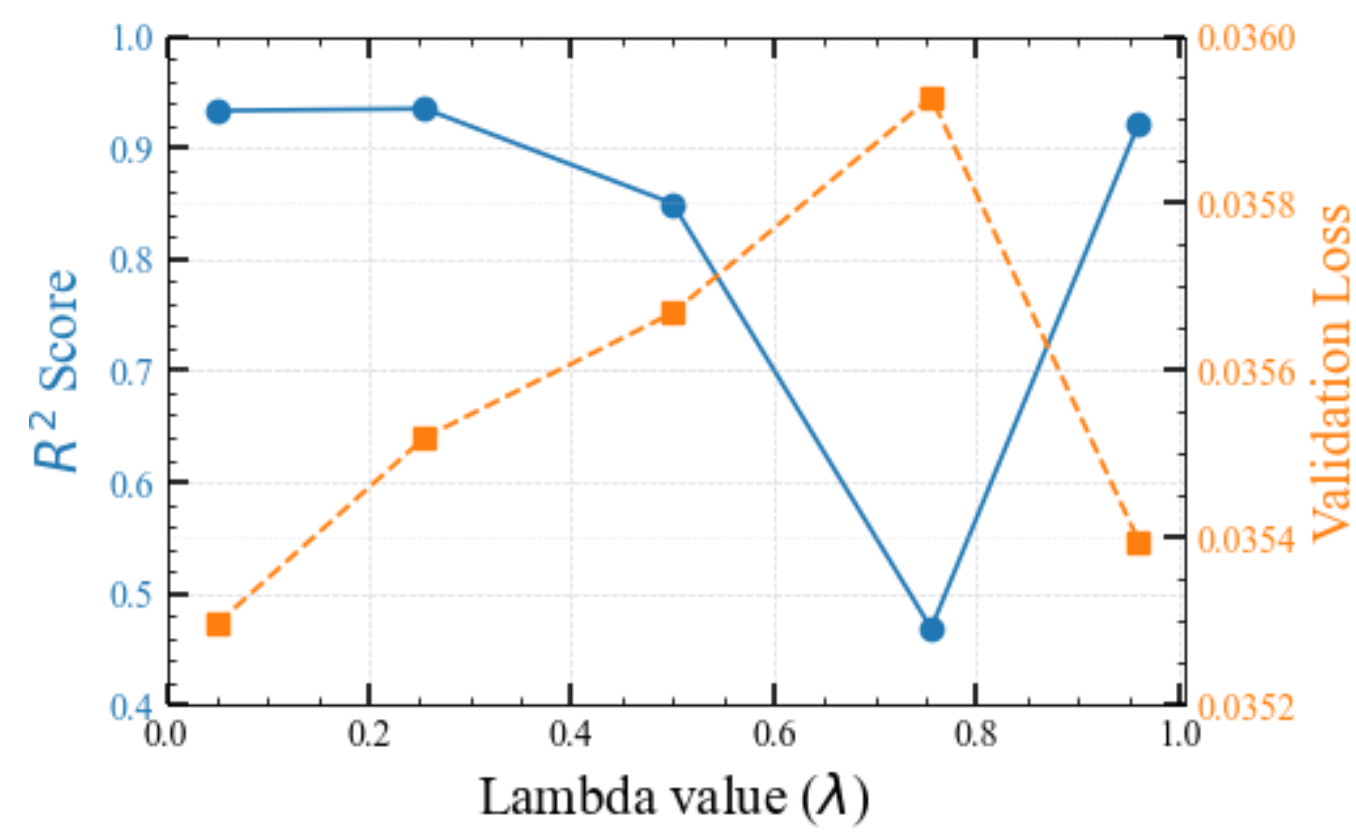

Fig. 10. Plot of $R^2$ score and validation score vs. λ.

Now, we will discuss the choice of $\lambda = 0.05$ in Section IV. We plot $R^2$ of the fitting as a function of $\lambda$ in Fig. 10. Choosing the optimal $\lambda$ is crucial because it balances the trade-off between empirical accuracy and adherence to physical constraints. For instance, if $\lambda$ is too small, the physics loss becomes negligible, causing the model to focus mainly on minimizing the mean squared error (MSE) loss. This leads to insufficient enforcement of the physical constraints. Conversely, if $\lambda$ is too high, the model overemphasizes these constraints at the expense of empirical accuracy, resulting in a poor fit on both the training and validation datasets and a diminished ability to generalize to new experimental data. In Table I, we can see this effect where $\lambda = 0.05$ has a few data points violating the physics constraint while for $\lambda = 0.96$, there are no physics violations at all. As for comparison in Table III, the $R^2$ score of using a PINN with $\lambda = 0.05$ in the linear scale was found to be 0.933 and when the model used a $\lambda = 0.96$, $R^2$ score drops to 0.92 which reflects the earlier statement. Ultimately, we chose $\lambda = 0.05$ because not only does it have the best $R^2$ score in Fig. 10, but also does it lie in a flat region of the curve where the $R^2$ score is relatively insensitive to small changes in $\lambda$. Mathematically, this is supported by the derivative $\frac{dR^2}{d\lambda}$, which is smaller in magnitude around the smaller $\lambda$. In contrast, for $\lambda = 0.96$, the $R^2$ score shows a steeper slope and greater variation. This suggests that although validation loss fluctuates across $\lambda$ values, the model at low $\lambda$ (e.g., 0.05) demonstrates greater robustness, as its $R^2$ score remains stable and unaffected by these fluctuations, whereas at higher $\lambda$, both $R^2$ and validation score become volatile. Therefore, $\lambda = 0.05$ is chosen to ensure a more robust AE-PINN model while still maintaining high accuracy and low physics violation model.

Finally, it should be noted that this study only attempts to find a set of PhuMob parameters that is closest to a correct one for an emerging material and device. It did not attempt to match individual IV curves which are noisy and have unknown sources of non-ideality (e.g., contact resistance, variations of doping, variations of thickness, and other unknown physics). However, the fact that TCAD IV simulations closely match the experimental IV at different temperatures and WF (high $R^2$) proved that the extracted parameters are reasonable. Particularly, the extracted values are similar to the manual one. The fact that the manual one has been validated in other applications (e.g., [4]) further justifies the usefulness of the extracted parameters.

## VII. Conclusions

We showed that it is possible to perform automatic TCAD mobility parameter calibration on novel materials such as $Ga_2O_3$ using ML. The machine is comprised of AE and PINN and is trained by TCAD-generated data. By experimentally measuring the device IV and feeding them into the machine, the quality of calibrated parameters is found to be as good as that obtained by a TCAD expert.

## VIII. Acknowledgment

This material is based upon work supported by the National Science Foundation under Grant No. 2046220.